\documentclass[conference]{IEEEtran}
\IEEEoverridecommandlockouts
\usepackage{cite}
\usepackage{amsmath,amssymb,amsfonts}
\usepackage{graphicx}
\usepackage{textcomp}
\usepackage{xcolor}
\usepackage{algorithm}
\usepackage{algpseudocode}
\usepackage{cite}
\usepackage{caption}
\usepackage{textcomp}
\usepackage[colorlinks=true, citecolor=customColor, linkcolor=blue, urlcolor=blue]{hyperref}
\definecolor{customColor}{RGB}{255, 0, 0} 

\usepackage{eso-pic} 

\newcommand\SideNote{%
    \AddToShipoutPictureBG*{%
        \put(10,60){\rotatebox{90}{\parbox{23cm}{\centering\scriptsize
        \textbf{Appeared in the 23rd IEEE/ACIS International Conference on Software Engineering, Management and Applications (SERA 2025)}
        }}}
    }
}

\SideNote

\def\BibTeX{{\rm B\kern-.05em{\sc i\kern-.025em b}\kern-.08em
    T\kern-.1667em\lower.7ex\hbox{E}\kern-.125emX}}
\begin{document}
\title{Evaluating BiLSTM and CNN+GRU Approaches for Human Activity Recognition Using WiFi CSI Data}
\author{
    \IEEEauthorblockN{
        Almustapha A. Wakili, 
        Babajide J. Asaju, and
        Woosub Jung}
    \IEEEauthorblockA{
        \textit{CIS Department, Towson University}\\
        Towson, Maryland, USA \\
        Email: \{awakili, basaju, woosubjung\}@towson.edu}
}

\maketitle

\begin{abstract}
This paper compares the performance of BiLSTM and CNN+GRU deep learning models for Human Activity Recognition (HAR) on two WiFi-based Channel State Information (CSI) datasets: UT-HAR and NTU-Fi HAR. The findings indicate that the CNN+GRU model has a higher accuracy on the UT-HAR dataset (95.20\%) thanks to its ability to extract spatial features. In contrast, the BiLSTM model performs better on the high-resolution NTU-Fi HAR dataset (92.05\%) by extracting long-term temporal dependencies more effectively. The findings strongly emphasize the critical role of dataset characteristics and preprocessing techniques in model performance improvement. We also show the real-world applicability of such models in applications like healthcare and intelligent home systems, highlighting their potential for unobtrusive activity recognition.
\end{abstract}

\begin{IEEEkeywords}
Human Activity Recognition; Deep Learning; WiFi Channel State Information (CSI);  Smart Cities. 
\end{IEEEkeywords}

\section{Introduction}
Human Activity Recognition (HAR) has become a critical area of research due to its vast applications in all areas of smart cities and healthcare, including security surveillance, smart home monitoring, and lifestyle management. For instance, HAR systems are instrumental in elderly care, enabling early detection of falls or health anomalies, and in smart homes, they enhance automation by understanding user behavior\cite{thakur2024human}. 

Traditionally, HAR relied on wearable sensors, such as accelerometers and gyroscopes, which, although effective, introduce limitations such as discomfort during long-term use and potential user resistance due to their intrusive nature \cite{kumar2024human}. WiFi-based systems, leveraging existing wireless infrastructure, have emerged as a promising alternative, enabling seamless, non-intrusive, and real-time activity recognition\cite{ahmad2024wifi}.

Deep learning has revolutionized HAR in recent years by providing models capable of processing complex, high-dimensional data streams from environmental sensors. Deep learning has been pivotal in advancing HAR, offering robust solutions capable of processing complex and high-dimensional data streams from environmental sensors \cite{ge2022contactless, ahmad2024wifi}. Bidirectional Long Short-Term Memory (BiLSTM) networks excel at capturing temporal dependencies in activity sequences, while Convolutional Neural Networks combined with Gated Recurrent Units (CNN+GRU) leverage the strengths of spatial and temporal feature extraction \cite{challa2022multibranch}. These models have shown promise in recognizing and classifying human activities, especially when combined with advanced preprocessing techniques that enhance WiFi Channel State Information (CSI) signals.

This study aims to comprehensively evaluate the BiLSTM and CNN+GRU models in HAR by utilizing two distinct WiFi-based CSI datasets: UT-HAR \cite{gringoli2019free} and NTU-Fi HAR \cite{yang2023sensefi}. By focusing on the effect of data pre-processing techniques, model architectures, and dataset characteristics, this study investigates how to improve HAR model performance regarding the accuracy and computational efficiency to make this model applicable for real-time in practical applications such as smart homes and healthcare environments. 

The rest of the paper is structured as follows: Section II discusses related work, Section III elaborates on the methodologies, Section IV presents experimental results, Section V discusses findings, and Section VI concludes with future directions.

\begin{figure}[htbp]
\centering
\includegraphics[width=0.5\textwidth]{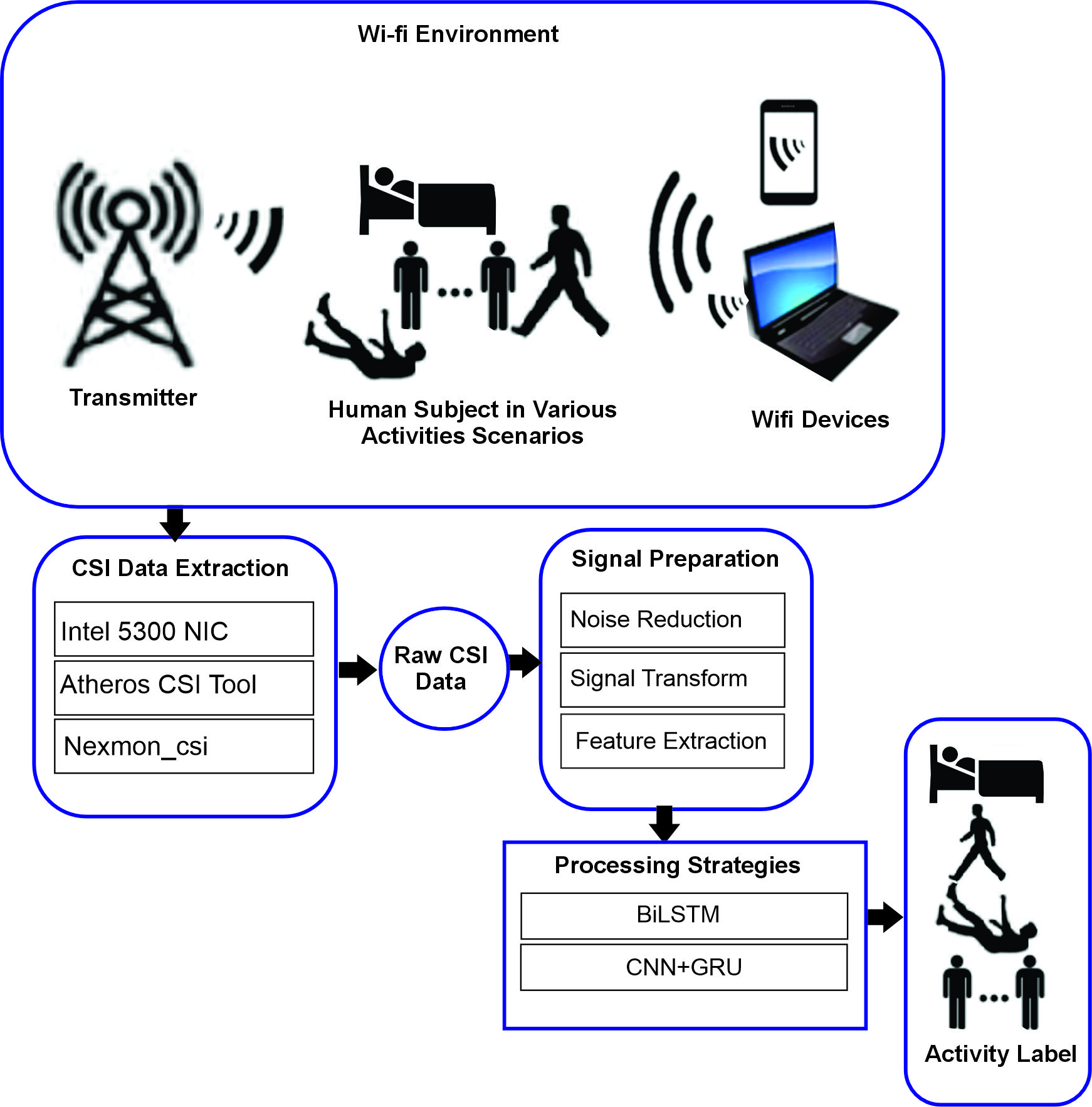}
\caption{WiFi-based Human Activity Recognition Pipeline.}
\label{fig:wifi_har_pipeline}
\end{figure}

The presented framework (Figure~\ref{fig:wifi_har_pipeline}) outlines the complete process of WiFi-based Human Activity Recognition (HAR). It begins with extracting Channel State Information (CSI) data using tools like Intel 5300 NIC, Atheros CSI Tool, and Nexmon CSI in a WiFi-enabled environment. The raw CSI data undergoes signal preparation, which includes noise reduction, signal transformation, and feature extraction, to improve its quality and usability. Finally, the processed signals are fed into BiLSTM and CNN+GRU models, which classify human activities into specific categories. This modular pipeline highlights the system's adaptability to different scenarios and potential applications in healthcare, smart homes, and other real-world settings.

\section{Related Works}

Human Activity Recognition (HAR) has gained widespread attention in recent years, driven by its transformative applications in healthcare, smart homes, and security systems. Traditional approaches primarily relied on wearable devices to monitor activities. While effective, these devices often raise concerns about user discomfort and practicality in daily life~\cite{shahverdi2024csi}. This limitation has motivated researchers to explore non-invasive solutions, particularly those leveraging WiFi signals for activity recognition.

\subsection{WiFi-Based Human Activity Recognition}
WiFi-based HAR offers a compelling alternative to sensor-dependent methods by utilizing Channel State Information (CSI) to capture the impact of human motion on wireless signals. Unlike wearable sensors, WiFi-based systems allow for seamless, device-free monitoring, which is especially appealing for applications in smart environments~\cite{ahmad2024wifi}. Device-free sensing approaches, which rely on ambient signals like WiFi, have proven particularly effective in scenarios where privacy is a concern. Compared to device-based sensing using wearable trackers, these methods eliminate the need for physical sensors on users, making them more suitable for healthcare and smart environments~\cite{shahverdi2024csi}.

Studies like Yang et al.~\cite{yang2023sensefi} have demonstrated the potential of WiFi-based HAR to detect activities with high precision, thanks to the fine-grained subcarrier-level information provided by CSI. Similarly, Ahmad et al.~\cite{ahmad2024wifi} explored the integration of WiFi signals with deep learning techniques to achieve precise activity classification, highlighting the non-invasive nature of RF-based sensing. However, the reliability of WiFi-based systems is often challenged by environmental noise, signal interference, and the complexity of real-world settings, such as non-line-of-sight (NLoS) scenarios.

Lightweight CNN architectures, as proposed by Zhang et al.~\cite{el2024csi}, have shown promise in addressing computational constraints while maintaining accuracy. Data augmentation techniques have also been employed to improve generalization across diverse environments. Advanced preprocessing pipelines incorporating denoising, phase correction, and data segmentation are fundamental to ensuring reliable model performance~\cite{9945513}. These preprocessing techniques enhance the clarity of CSI data and mitigate the impact of environmental noise, a critical step in developing robust HAR systems.

\subsection{Deep Learning Approaches in HAR}
The rise of deep learning has significantly advanced the capabilities of HAR systems, providing tools to analyze complex data patterns and uncover subtle features within activity datasets. Among the most prominent models are Bidirectional Long Short-Term Memory (BiLSTM) networks, which excel at capturing temporal dependencies. For example, Li et al.~\cite{li2019bi} demonstrated the effectiveness of BiLSTM in recognizing sequential activity patterns by leveraging its ability to process data in both forward and backward temporal directions. This bidirectional nature allows the model to capture more nuanced transitions between activities.

Hybrid architectures, such as those combining Convolutional Neural Networks (CNNs) with Gated Recurrent Units (GRUs), have also emerged as powerful alternatives. By integrating spatial feature extraction through CNNs and temporal sequence modeling through GRUs, these models offer a balanced approach to processing activity data. Studies like Kim et al.~\cite{kim2021wearable} and Shahverdi et al.~\cite{shahverdi2024csi} highlight the efficiency of CNN+GRU architectures, particularly in datasets with lower temporal resolution, where spatial patterns play a more critical role. However, the trade-offs between computational efficiency and temporal modeling remain an ongoing challenge~\cite{9945368}.

Emerging technologies, including transformer-based models, offer promising solutions to these challenges. These architectures, known for their ability to model long-range dependencies, are beginning to be adapted for HAR tasks~\cite{kumar2024human}. Additionally, multimodal approaches integrating visual and non-visual modalities have been explored to enhance activity recognition in complex settings, as demonstrated by Bouchabou et al.~\cite{bouchabou2021survey}.

\subsection{Preprocessing for CSI Data}
The importance of preprocessing cannot be overstated in WiFi-based HAR, as raw CSI data is often noisy and susceptible to interference. Preprocessing techniques such as denoising, normalization, and sliding window segmentation significantly influence model accuracy~\cite{yousefi2017survey}. Fourier and wavelet transforms are widely used to extract frequency-domain features that reveal periodic patterns associated with human activities~\cite{yang2023sensefi}. High-pass filtering is another common technique employed to eliminate low-frequency noise, enhancing signal quality.

Recent work by Yousefi et al.~\cite{yousefi2017survey} introduced Body-coordinate Velocity Profiles (BVP) to reduce environmental dependencies by focusing on motion-specific characteristics of CSI. However, challenges persist even with these advanced methods. While effective, techniques like sliding window segmentation often overlook fine-grained temporal variations, which can be critical for distinguishing between similar activities. Advanced architectures, such as attention mechanisms and transformers, may provide a pathway to overcoming these limitations by improving temporal feature extraction~\cite{kumar2024human}.

\subsection{Current Challenges and Emerging Trends}
Despite significant advancements, HAR research faces limited dataset diversity, scalability, and real-time performance. Current datasets often lack the variability needed to generalize models across different environments, including diverse layouts, overlapping activities, and dynamic noise sources~\cite{ahmad2024wifi}. Lightweight models, such as those explored by Zhang et al.~\cite{el2024csi}, and large-scale annotated datasets have been identified as critical for advancing the field.

As highlighted by Kumar et al.~\cite{kumar2024human}, transformer architectures and attention mechanisms offer promising solutions by prioritizing relevant features and discarding irrelevant noise. These innovations could significantly enhance the interpretability and robustness of HAR systems. Future research should also address trade-offs between accuracy and computational cost, particularly for resource-constrained devices~\cite{9945368}.

Interdisciplinary collaborations are increasingly considered vital to advancing HAR research. Integrating domain knowledge from healthcare, ergonomics, and urban design could lead to more adaptable and user-centric solutions, ensuring that HAR technologies meet the demands of practical, real-world applications. This review establishes the foundation for our study, focusing on leveraging deep learning techniques, particularly BiLSTM and CNN+GRU models, to enhance WiFi-based HAR systems for diverse applications.

\section{Methodology}
This study evaluates the performance of BiLSTM and CNN+GRU models on two WiFi-based CSI datasets: NTU\_Fi\_HAR and UT\_HAR. The methodology includes dataset preprocessing, model implementation, training strategies, and computational efficiency analysis.

\subsection{Dataset Overview}
\subsubsection{NTU\_Fi\_HAR Dataset}
The NTU\_Fi\_HAR dataset, collected using the Atheros CSI Tool, features high-resolution CSI data with 114 subcarriers per antenna pair and captures a variety of human activities in different scenarios \cite{yang2023sensefi}. This dataset includes amplitude and phase information, enabling a comprehensive analysis of the CSI data's spatial and temporal features.

Table \ref{table:ntu_fi_har_activities} shows the distribution of training and validation instances across the six activity classes included in the dataset.

\begin{table}[ht]
\centering
\renewcommand{\arraystretch}{1.2}
\setlength{\tabcolsep}{8pt}
\begin{tabular}{|c|c|c|}
\hline
\textbf{Activity} & \textbf{Training Instances} & \textbf{Validation Instances} \\ \hline
Clean & 156 & 44 \\ 
Fall & 156 & 44 \\ 
Run & 156 & 44 \\ 
Walk & 156 & 44 \\ 
Jump & 156 & 44 \\ 
Circle & 156 & 44 \\ 
\hline
\end{tabular}
\caption{Distribution of activities and the number of instances in the NTU\_Fi\_HAR dataset.}
\label{table:ntu_fi_har_activities}
\end{table}

\subsubsection{UT\_HAR Dataset}
The UT\_HAR dataset, collected using the Intel 5300 NIC, is designed for HAR tasks and features continuously recorded CSI data. Unlike NTU\_Fi\_HAR, this dataset lacks explicit activity segmentation. Therefore, preprocessing steps, such as applying a sliding window technique, are required to segment the data into meaningful samples \cite{gringoli2019free}.

Table \ref{table:ut_har_classes} presents the distribution of training, validation, and testing instances for the six activity classes included in the dataset.

\begin{table}[ht]
\centering
\renewcommand{\arraystretch}{1.2}
\setlength{\tabcolsep}{8pt}
\resizebox{0.5\textwidth}{!}{%
\begin{tabular}{|c|c|c|c|}
\hline
\textbf{Class} & \textbf{Training Instances} & \textbf{Validation Instances} & \textbf{Testing Instances} \\ \hline
Class 0 & 3977 & 496 & 500 \\ 
Class 1 & 3977 & 496 & 500 \\ 
Class 2 & 3977 & 496 & 500 \\ 
Class 3 & 3977 & 496 & 500 \\ 
Class 4 & 3977 & 496 & 500 \\ 
Class 5 & 3977 & 496 & 500 \\ 
\hline
\end{tabular}
}
\caption{Distribution of classes and the number of instances in the UT\_HAR dataset.}
\label{table:ut_har_classes}
\end{table}

\subsection{Preprocessing}
Effective preprocessing of raw CSI data is crucial for achieving high model performance. This study employs the following preprocessing steps tailored to the characteristics of the NTU\_Fi\_HAR and UT\_HAR datasets:

\begin{itemize}
    \item \textbf{Noise Reduction:} High-pass filters are applied to eliminate low-frequency noise, enhancing signal clarity for activity recognition.
    \item \textbf{Normalization:} CSI amplitude and phase data are scaled uniformly, ensuring sample consistency and reducing data biases.
    \item \textbf{Feature Extraction:} Fourier and wavelet transforms capture essential frequency and time-domain features, critical for spatial and temporal analysis.
\end{itemize}

Table \ref{table:preprocessing_summary} summarizes the preprocessing pipeline, highlighting the unique adaptations for each dataset.

\begin{table}[ht]
\centering
\renewcommand{\arraystretch}{1.2}
\setlength{\tabcolsep}{8pt}
\resizebox{0.5\textwidth}{!}{%
\begin{tabular}{|c|c|c|}
\hline
\textbf{Preprocessing Step} & \textbf{NTU\_Fi\_HAR} & \textbf{UT\_HAR} \\ \hline
Noise Reduction & High-pass filtering & High-pass filtering \\ 
Normalization & Amplitude and phase scaling & Amplitude scaling \\ 
Feature Extraction & Doppler profiles, Fourier transforms & Sliding window segmentation \\ 
\hline
\end{tabular}
}
\caption{Comparison of preprocessing steps for NTU\_Fi\_HAR and UT\_HAR datasets.}
\label{table:preprocessing_summary}
\end{table}
 
\begin{figure*}[htbp]
\centering
\includegraphics[width=0.91\textwidth]{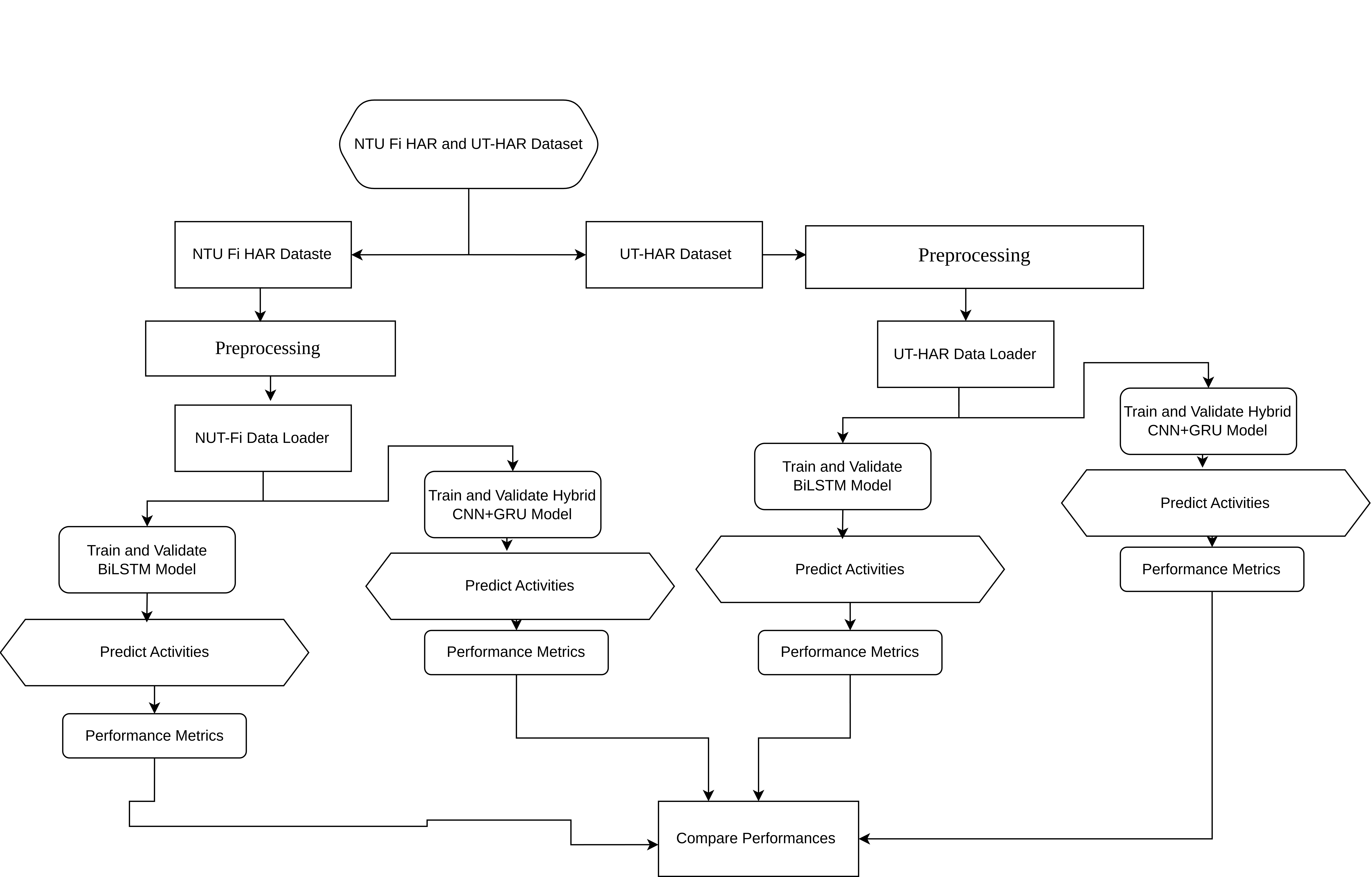}
\caption{Workflow for Training and Evaluating BiLSTM and CNN+GRU Models on NTU-Fi HAR and UT-HAR Datasets.}
\label{fig:har_workflow}
\end{figure*}
\color{black}

The workflow illustrated in Figure~\ref{fig:har_workflow} provides a detailed overview of the experimental methodology used in this study. The process starts with the NTU-Fi HAR and UT-HAR datasets, each undergoing specific preprocessing steps such as noise reduction, feature extraction, and segmentation. Preprocessed data is loaded into separate pipelines for training and validation using BiLSTM and CNN+GRU models. These models predict human activities based on the processed data, and their performance metrics are computed for evaluation. Finally, the performances of both models are compared to understand their relative strengths and suitability for different datasets and scenarios.

\subsection{Model Implementation}

\subsubsection{BiLSTM Model}
BiLSTM is designed to capture long-term dependencies. The model includes three bidirectional LSTM layers, each with 128 units, followed by a dense layer with softmax activation.

\subsubsection{CNN+GRU Model}
\begin{figure*}[htbp]
\centerline{\includegraphics[width=0.79\textwidth]{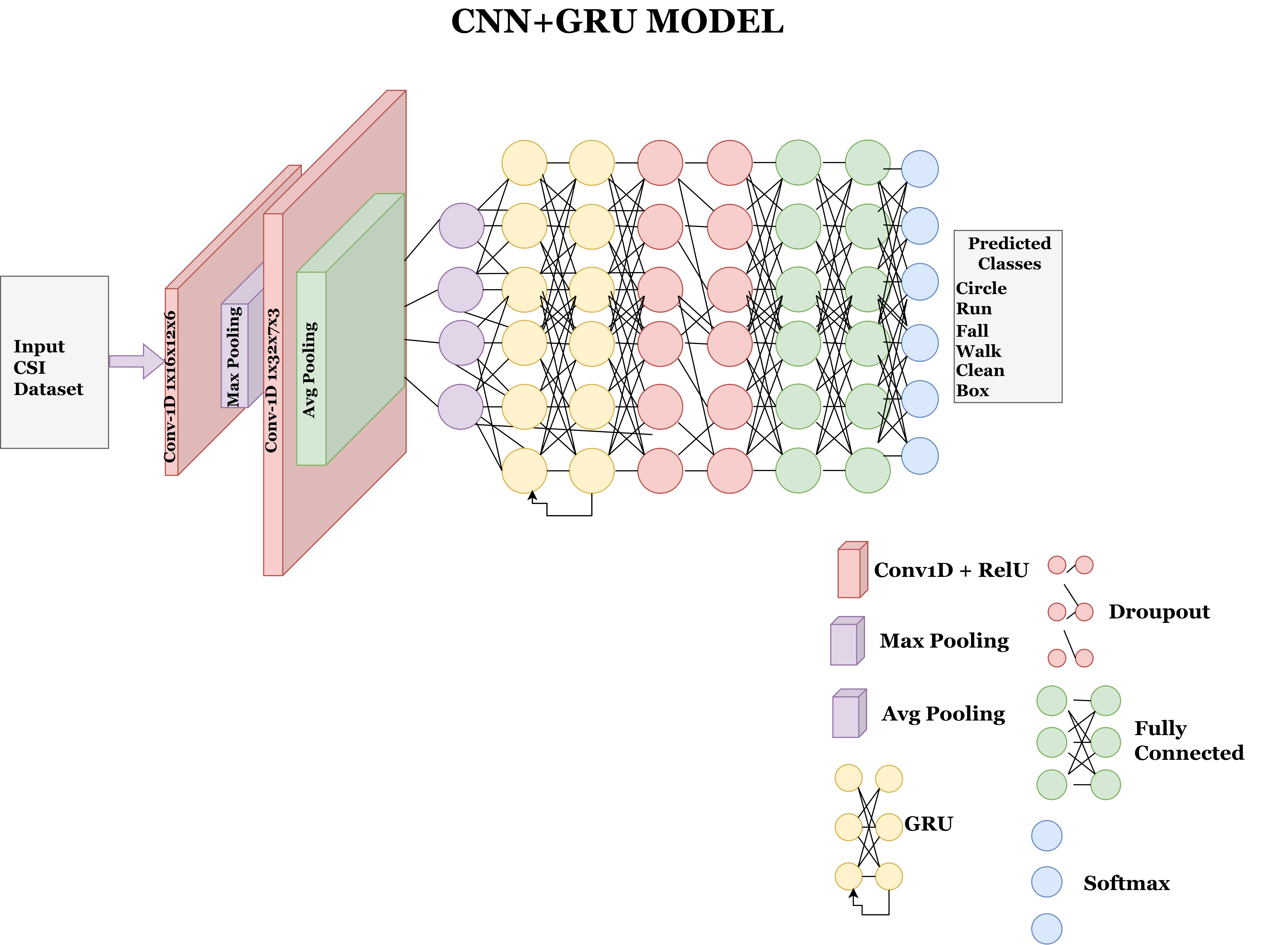}}
\caption{CNN+GRU Model Architecture. This hybrid architecture integrates convolutional layers with GRU layers to enhance spatiotemporal analysis.}
\label{fig:cnn_gru_architecture}
\end{figure*}

The CNN+GRU model combines convolutional layers for spatial feature extraction with GRU layers for temporal sequence analysis. This hybrid architecture enhances the model's ability to handle the spatiotemporal characteristics of CSI data. Figure \ref{fig:cnn_gru_architecture} shows the CNN+GRU architecture.

\subsection{Training and Computational Efficiency}
Both models are trained using the Adam optimizer with a learning rate 0.001 and early stopping to prevent overfitting. Table~\ref{table:computational_efficiency} compares the inference time and memory usage, highlighting the computational advantages of CNN+GRU.

\begin{table}[ht]
\centering
\begin{tabular}{|c|c|c|}
\hline
Model & Inference Time (milliseconds) & Memory Usage (MB) \\ \hline
BiLSTM & 120 & 250 \\ 
CNN+GRU & 90 & 150 \\ 
\hline
\end{tabular}
\caption{Comparison of Computational Efficiency. This table shows the inference time and memory usage for the BiLSTM and CNN+GRU models.}
\label{table:computational_efficiency}
\end{table}

\section{Results and Evaluation}
This section presents the experimental results and provides a comparative analysis of the performance of the CNN+GRU and BiLSTM models on the NTU-Fi HAR and UT-HAR datasets. The results highlight distinct trends in model performance, reflecting the influence of dataset characteristics and architectural strengths. The CNN+GRU model consistently demonstrated superior performance on the UT-HAR dataset, primarily due to its advanced spatial feature extraction capabilities. Conversely, the BiLSTM model excelled on the NTU-Fi HAR dataset, effectively leveraging its ability to capture long-term temporal dependencies from high-resolution CSI data.

\subsection{Performance on NTU\_Fi\_HAR Dataset}
The NTU\_Fi\_HAR dataset, characterized by its high-resolution CSI data, provided a robust platform for evaluating both temporal and spatial feature extraction capabilities. Table~\ref{table:ntu_fi_har_metrics} summarizes the overall performance metrics for CNN+GRU and BiLSTM models. The CNN+GRU model achieved the highest overall accuracy, precision, recall, and F1-score, highlighting its versatility. However, the BiLSTM model also performed strongly, particularly in activities that require intricate temporal analysis.

\begin{table}[ht]
\centering
\renewcommand{\arraystretch}{1.2}
\setlength{\tabcolsep}{7pt}
\resizebox{0.45\textwidth}{!}{%
\begin{tabular}{|c|c|c|c|c|}
\hline
\textbf{Model} & \textbf{Accuracy(\%)} & \textbf{Precision(\%)} & \textbf{Recall(\%)} & \textbf{F1-Score(\%)} \\ \hline
CNN+GRU & 99.24 & 99.24 & 99.24 & 99.24 \\ 
BiLSTM & 92.05 & 92.83 & 92.05 & 92.03 \\ 
\hline
\end{tabular}
}
\caption{Overall performance metrics for CNN+GRU and BiLSTM models on the NTU\_Fi\_HAR dataset.}
\label{table:ntu_fi_har_metrics}
\end{table}

Table~\ref{table:ntu_fi_har_classwise} provides a detailed class-wise performance comparison. The CNN+GRU model consistently outperformed BiLSTM across all classes, with notable precision and recall for activities such as "Run" and "Walk." The BiLSTM model, while performing adequately overall, exhibited minor misclassifications in more complex activities like "Run" and "Box," possibly due to the intricate temporal patterns involved.

\begin{table}[ht]
\centering
\renewcommand{\arraystretch}{1.2}
\setlength{\tabcolsep}{6pt}
\resizebox{0.4\textwidth}{!}{%
\begin{tabular}{|c|c|c|c|c|c|c|}
\hline
\textbf{Class} & \multicolumn{3}{|c|}{\textbf{CNN+GRU}} & \multicolumn{3}{|c|}{\textbf{BiLSTM}} \\ \hline
\textbf{} & \textbf{Precision} & \textbf{Recall} & \textbf{F1} & \textbf{Precision} & \textbf{Recall} & \textbf{F1} \\ \hline
Clean & 98 & 100 & 99 & 100 & 93 & 96 \\ 
Fall & 100 & 100 & 100 & 100 & 100 & 100 \\ 
Run & 100 & 98 & 99 & 82 & 91 & 86 \\ 
Walk & 100 & 98 & 99 & 86 & 100 & 93 \\ 
Box & 98 & 100 & 99 & 89 & 77 & 83 \\ 
Circle & 100 & 100 & 100 & 98 & 91 & 94 \\ 
\hline
\end{tabular}
}
\caption{Class-wise performance metrics (\%) for CNN+GRU and BiLSTM models on the NTU\_Fi\_HAR dataset.}
\label{table:ntu_fi_har_classwise}
\end{table}

\begin{figure}[ht]
\centering
\includegraphics[width=0.5\textwidth]{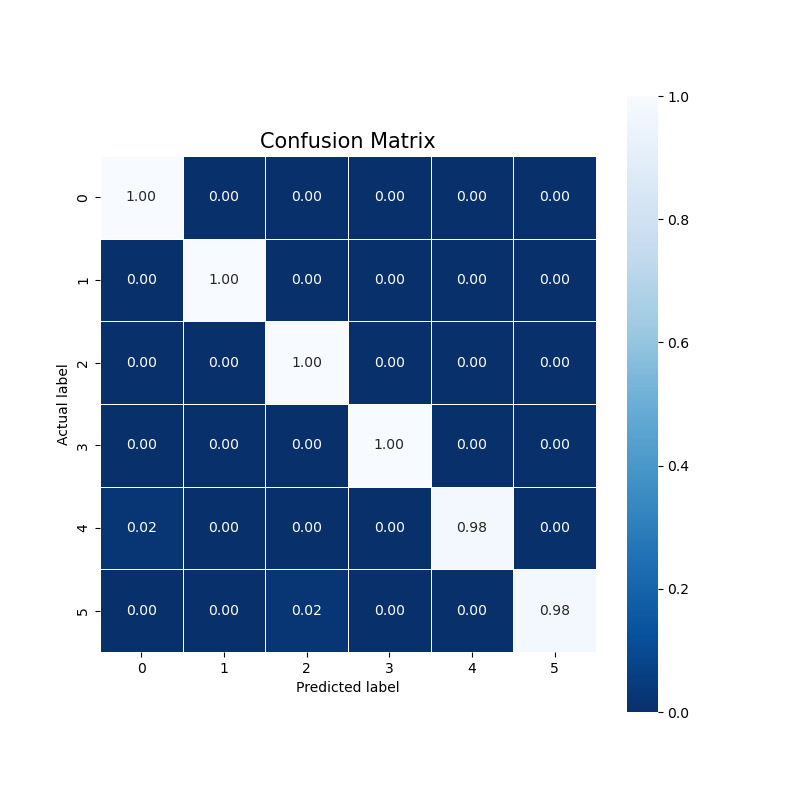}
\caption{Confusion matrix for CNN+GRU model on NTU-Fi HAR dataset, demonstrating high classification accuracy across all activities. Minor misclassifications are evident but do not significantly impact performance.}
\label{fig:ntu_fi_har_cnn_gru_confusion}
\end{figure}

\begin{figure}[ht]
\centering
\includegraphics[width=0.5\textwidth]{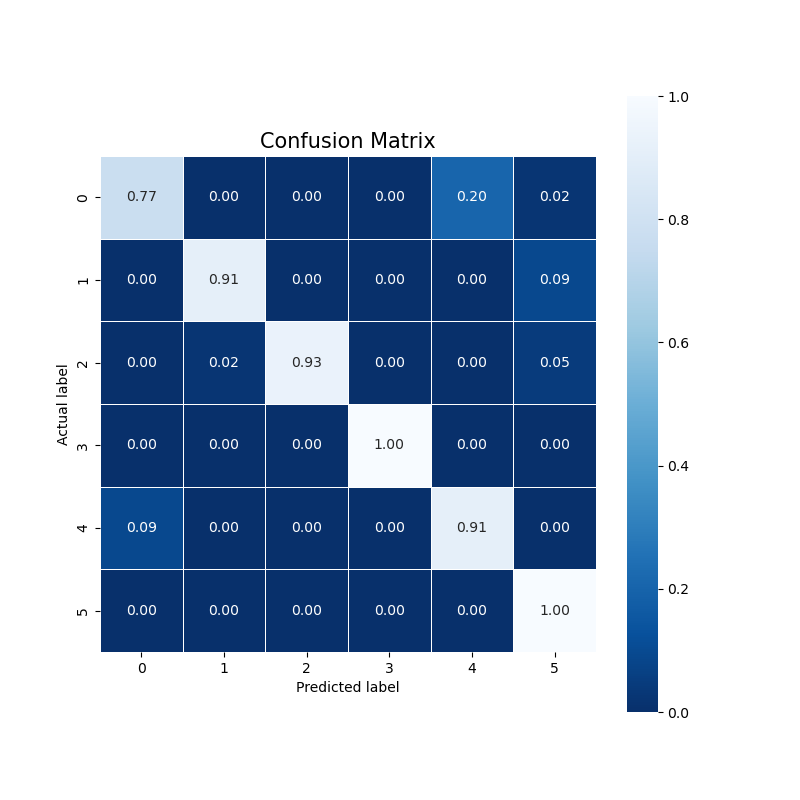}
\caption{Confusion matrix for BiLSTM model on NTU-Fi HAR dataset, showcasing strong classification performance with minor misclassifications in complex activities.}
\label{fig:ntu_fi_har_bilstm_confusion}
\end{figure}

Figure~\ref{fig:ntu_fi_har_cnn_gru_confusion} illustrates the confusion matrix for the CNN+GRU model on the NTU-Fi HAR dataset. The matrix reflects near-perfect classification accuracy across all activity classes, with only minor misclassifications observed in activities such as "Box." Similarly, Figure~\ref{fig:ntu_fi_har_bilstm_confusion} shows the confusion matrix for the BiLSTM model, which, while performing well overall, struggled slightly in distinguishing between similar activities like "Run" and "Box," likely due to overlapping temporal patterns.

\subsection{Performance on UT\_HAR Dataset}
The UT-HAR dataset, characterized by lower resolution and limited temporal details, presented unique challenges for both models. Table~\ref{table:ut_har_metrics} summarizes the overall performance, with the CNN+GRU model outperforming BiLSTM in accuracy, precision, recall, and F1-score. The CNN+GRU model's superior ability to extract spatial features was instrumental in achieving higher accuracy, particularly given the reduced temporal resolution of the UT-HAR dataset.

\begin{table}[ht]
\centering
\renewcommand{\arraystretch}{1.2}
\setlength{\tabcolsep}{8pt}
\resizebox{0.45\textwidth}{!}{%
\begin{tabular}{|c|c|c|c|c|}
\hline
\textbf{Model} & \textbf{Accuracy(\%)} & \textbf{Precision(\%)} & \textbf{Recall(\%)} & \textbf{F1-Score(\%)} \\ \hline
CNN+GRU & 95.20 & 94.32 & 93.70 & 93.72 \\ 
BiLSTM & 66.80 & 64.35 & 59.29 & 59.26 \\ 
\hline
\end{tabular}
}
\caption{Overall performance metrics for CNN+GRU and BiLSTM models on the UT-HAR dataset.}
\label{table:ut_har_metrics}
\end{table}

Class-wise performance metrics are detailed in Table~\ref{table:ut_har_classwise}. While the CNN+GRU model performed consistently across all classes, the BiLSTM model encountered significant challenges, particularly in activities such as "Walk" and "Box." These results underline the limitations of the BiLSTM model in handling datasets with constrained temporal information.

\begin{table}[ht]
\centering
\renewcommand{\arraystretch}{1.2}
\setlength{\tabcolsep}{6pt}
\begin{tabular}{|c|c|c|c|c|c|c|}
\hline
\textbf{Class} & \multicolumn{3}{|c|}{\textbf{CNN+GRU}} & \multicolumn{3}{|c|}{\textbf{BiLSTM}} \\ \hline
\textbf{} & \textbf{Precision} & \textbf{Recall} & \textbf{F1} & \textbf{Precision} & \textbf{Recall} & \textbf{F1} \\ \hline
Clean & 96 & 96 & 96 & 73 & 85 & 78 \\ 
Fall & 91 & 100 & 95 & 65 & 52 & 58 \\ 
Run & 98 & 99 & 98 & 79 & 70 & 74 \\ 
Walk & 87 & 85 & 86 & 38 & 28 & 32 \\ 
Box & 100 & 91 & 95 & 55 & 56 & 55 \\ 
Circle & 96 & 98 & 97 & 64 & 82 & 72 \\ 
\hline
\end{tabular}
\caption{Class-wise performance metrics (\%) for CNN+GRU and BiLSTM models on the UT-HAR dataset.}
\label{table:ut_har_classwise}
\end{table}

\begin{figure}[ht]
\centering
\includegraphics[width=0.45\textwidth]{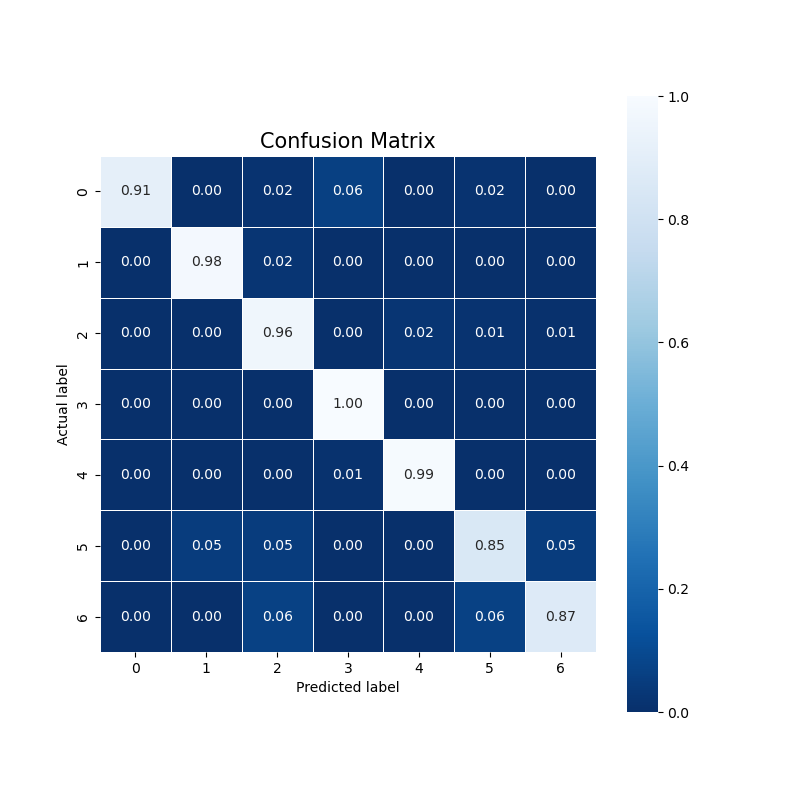}
\caption{Confusion matrix for CNN+GRU model on UT-HAR dataset, indicating strong spatial feature extraction capabilities with minimal misclassifications.}
\label{fig:ut_har_cnn_gru_confusion}
\end{figure}

\begin{figure}[ht]
\centering
\includegraphics[width=0.45\textwidth]{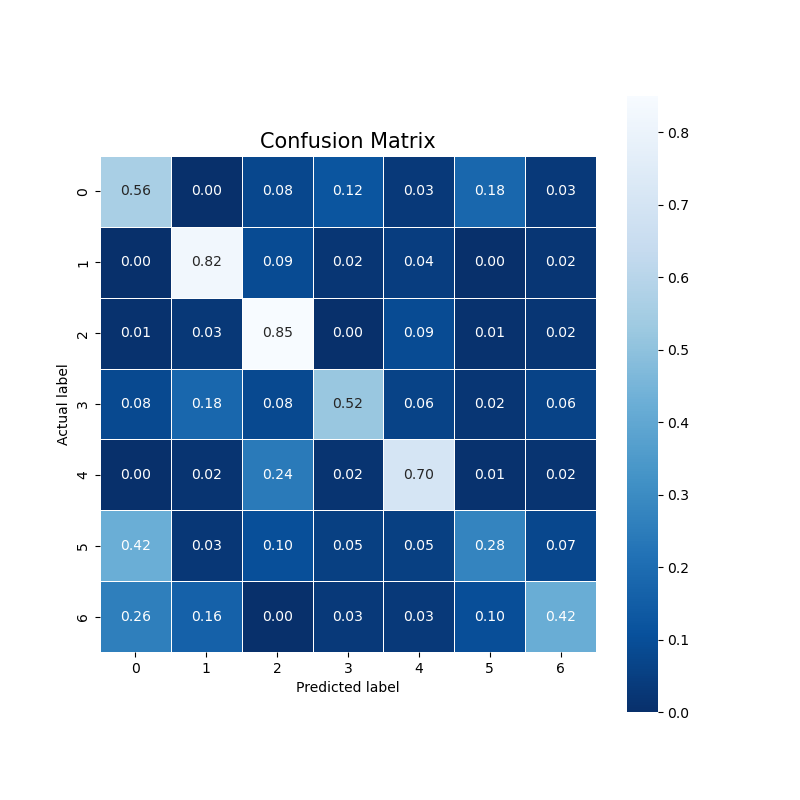}
\caption{Confusion matrix for BiLSTM model on UT-HAR dataset, showing substantial misclassifications in activities with less distinct temporal dependencies.}
\label{fig:ut_har_bilstm_confusion}
\end{figure}

The confusion matrix for the CNN+GRU model, presented in Figure~\ref{fig:ut_har_cnn_gru_confusion}, highlights its robustness in classifying most activities accurately, with only minor misclassifications observed in activities such as "Clean" and "Fall." Figure~\ref{fig:ut_har_bilstm_confusion} illustrates the confusion matrix for the BiLSTM model, which reveals substantial misclassifications in activities like "Walk" and "Box," underscoring its limited suitability for datasets with restricted temporal granularity.

\subsection{Summary}
In summary, the CNN+GRU model outperformed BiLSTM across both datasets, particularly excelling on the UT-HAR dataset due to its advanced spatial feature extraction capabilities. While the BiLSTM model demonstrated strengths in capturing temporal dependencies on the NTU-Fi HAR dataset, its overall performance remained inferior to the CNN+GRU model, particularly in scenarios requiring efficient handling of spatial features.

\subsection{Discussion}
The experimental results underscore the critical importance of aligning model architectures with dataset characteristics to achieve optimal Human Activity Recognition (HAR) performance. The CNN+GRU model consistently outperformed the BiLSTM model on the UT-HAR dataset, highlighting its superior ability to extract spatial features from segmented and lower-resolution CSI data. Conversely, the BiLSTM model excelled on the NTU-Fi HAR dataset due to its ability to capture long-term temporal dependencies inherent in high-resolution CSI data.

The observed performance differences between the NTU\_Fi\_HAR and UT\_HAR datasets stem from several critical factors, including the resolution and quality of the collected CSI data, the applied preprocessing techniques, and the inherent properties of the datasets. These findings provide valuable insights into the interplay between dataset characteristics and model architecture in HAR applications.

\subsubsection{Key Contributors to Performance}
\paragraph{CSI Collection Tools.} 
The quality and resolution of the CSI data play a pivotal role in model performance. The NTU\_Fi\_HAR dataset, collected using the Atheros CSI Tool, offers high-resolution data with 114 subcarriers per antenna pair. This high granularity enables models to capture subtle patterns associated with human activities, benefiting architectures like BiLSTM that excel in temporal feature extraction. In contrast, the UT\_HAR dataset, collected using the Intel 5300 NIC, provides lower-resolution data with fewer subcarriers, which limits the ability of temporal models like BiLSTM to achieve optimal performance. This limitation, however, is better managed by the CNN+GRU model, which effectively extracts spatial features even from lower-resolution data.

\paragraph{Preprocessing Techniques.}
The preprocessing pipeline significantly influences model performance by enhancing the quality and interpretability of raw CSI data. The NTU\_Fi\_HAR dataset benefited from comprehensive preprocessing, including:
\begin{itemize}
    \item \textbf{Noise Reduction:} High-pass filters removed low-frequency noise, improving signal clarity.
    \item \textbf{Normalization:} Scaling of amplitude and phase data ensured consistency across samples.
    \item \textbf{Feature Extraction:} Fourier and wavelet transform captured critical frequency and time-domain features.
    \item \textbf{Processed Doppler Representation:} Advanced techniques, such as Body-coordinate Velocity Profiles (BVP), reduced environmental dependencies while emphasizing human motion.
\end{itemize}
In contrast, preprocessing for the UT\_HAR dataset was more basic, primarily involving sliding window segmentation and statistical feature extraction. While effective, this limited preprocessing pipeline constrained the BiLSTM model's ability to utilize the temporal aspects of the data fully.

\paragraph{Dataset Characteristics.}
The NTU\_Fi\_HAR dataset's high resolution aligns well with BiLSTM's ability to model detailed temporal patterns, resulting in superior performance for complex, time-sensitive activities. On the other hand, the lower resolution of the UT\_HAR dataset posed challenges for BiLSTM, while the CNN+GRU model adapted well by leveraging its combined spatial and temporal feature extraction capabilities.

\paragraph{Temporal and Spatial Feature Extraction.}
The CNN+GRU model demonstrated its effectiveness in scenarios where spatial patterns dominate, as seen in the UT-HAR dataset. Its convolutional layers efficiently captured spatial features, while the GRU layers modeled temporal dynamics. In contrast, the BiLSTM model excelled in capturing temporal dependencies in the NTU-Fi HAR dataset, where the high resolution provided sufficient temporal granularity. These results underscore the importance of matching model architecture to the specific demands of the dataset.

\subsubsection{Implications of the Study}
The findings highlight the potential of hybrid deep learning architectures, such as CNN+GRU, for real-time HAR applications, particularly in resource-constrained environments. High-resolution datasets like NTU\_Fi\_HAR benefit from advanced temporal modeling techniques, while lower-resolution datasets like UT\_HAR require robust spatial feature extraction. These results emphasize the importance of high-quality data collection tools and preprocessing pipelines in enhancing model performance.

\section{Limitations and Future Directions}
Despite the promising results, this study has certain limitations that present opportunities for future research:
\begin{itemize}
    \item \textbf{Dataset Diversity:} The datasets utilized in this study primarily represent controlled environments. Extending the evaluation to more diverse and dynamic scenarios, such as outdoor settings or crowded environments, would improve the generalizability of the findings.
    \item \textbf{Computational Constraints:} While the BiLSTM model demonstrated strong performance on high-resolution data, its higher inference times and memory usage limit its applicability in resource-constrained devices. Future work could focus on optimizing BiLSTM for such environments.
    \item \textbf{Temporal Variability:} Sliding window segmentation, while effective, may overlook finer temporal variations. Advanced temporal modeling techniques, such as attention mechanisms or transformers, could enhance temporal feature extraction and improve model accuracy.
\end{itemize}
Future research should explore integrating attention mechanisms and transformer architectures to improve temporal and spatial feature extraction. Additionally, testing these models on large-scale, real-world datasets would validate their scalability and robustness for practical applications.

\section{Conclusion}
This study compares the BiLSTM and CNN+GRU architectures on UT‐HAR and NTU‐Fi HAR datasets, emphasizing how dataset characteristics and preprocessing workflows shape the performance of deep learning–based HAR models. The CNN+GRU model excels in scenarios where spatial feature extraction is paramount (UT‐HAR), while the BiLSTM model demonstrates superior performance with high‐resolution temporal data (NTU‐Fi HAR). These findings underscore the importance of aligning model choice with data characteristics in real‐world deployments, such as smart homes and healthcare monitoring systems. This research provides practical insights into choosing and tailoring advanced deep learning models for robust, scalable, and efficient HAR applications.ns.

\bibliographystyle{ieeetr}
\bibliography{ref}

\end{document}